# Modeling Events with Cascades of Poisson Processes

**Aleksandr Simma**
EECS Department
University of California, Berkeley
alex@asimma.com

**Michael I. Jordan**
Depts. of EECS and Statistics
University of California, Berkeley
jordan@cs.berkeley.edu

## Abstract

We present a probabilistic model of events in continuous time in which each event triggers a Poisson process of successor events. The ensemble of observed events is thereby modeled as a superposition of Poisson processes. Efficient inference is feasible under this model with an EM algorithm. Moreover, the EM algorithm can be implemented as a distributed algorithm, permitting the model to be applied to very large datasets. We apply these techniques to the modeling of Twitter messages and the revision history of Wikipedia.

## 1 Introduction

Real-life observations are often naturally represented by *events*—bundles of features that occur at a particular moment in time. Events are generally non-independent: one event may cause others to occur. Given observations of events, we wish to produce a probabilistic model that can be used not only for prediction and parameter estimation, but also for identifying structure and relationships in the data generating process.

We present an approach for building probabilistic models for collections of events in which each event induces a Poisson process of triggered events. This approach lends itself to efficient inference with an EM algorithm that can be distributed across computing clusters and thereby applied to massive datasets. We present two case studies, the first involving a collection of Twitter messages on financial data, and the second focusing on the revision history of Wikipedia. The latter example is a particularly large-scale problem; the data consist of billions of potential interactions among events.

Our approach is based on a continuous-time formalism. There have been a relatively small number of machine learning papers focused on continuous-time graphical models; examples include the "Poisson networks" of Rajaram et al. [2005] and the "continuous-time Bayesian networks" described in Nodelman et al. [2002, 2005]. These approaches differ from ours in that they assume a small set of possible event labels and do not directly apply to structured label spaces. A more flexible approach has been presented by Wingate et al. [2009] who define a nonparametric Bayesian model with latent events and causal structure. This work differs from ours in several ways, most importantly in that it is a discrete-time model that allows for interaction only between adjacent time steps. Finally, this work is an extension and generalization of the "continuous-time noisy-or" presented in Simma et al. [2008].

There is also a large literature in statistics on point process modeling that provides a context for our work. A specific connection is that the fundamental stochastic process in our model is known in statistics as a "mutually self-exciting point process" [Hawkes, 1971]. There are also connections to applications in seismology, notably the "Epidemic Type Aftershock-Sequences" framework of Ogata [1988], which involves a model similar to ours that is applied to earthquake prediction.

## 2 Modeling Events with Poisson and Cox Processes

Our representation of collections of events is based on the formalism of *marked point processes*. Let each event be represented as a pair $(t, x) \in \mathcal{R}^+ \times \mathcal{F}$, where $t$ is the timestamp and $x$ the associated features taking values in a feature space. A dataset is a sequence of observations $(t, x) \in \mathcal{R}^+ \times \mathcal{F}$. We use $D_{a:b}$ to denote the events occuring between times $a$ and $b$.

Within the framework of marked point processes, we have several modeling issues to address: 1) how many events occur? 2) when do events occur? 3) what features they possess? A classical approach to answering these questions proceeds as follows: 1) the number is distributed Poisson($\alpha$), 2) the timestamps associated with event are independent and identically distributed (iid) from a fixed distribution, 3) the features are drawn independently from a fixed distribution $g$:

$$\begin{aligned} \text{the density} : f\left(t, x\right) &= f_\theta = T \cdot \alpha \cdot h(t) g(x) \\ \text{the data} : D_{0:T} &\sim \text{PP}\left(f\right), \end{aligned}$$

where $\alpha$ is the average occurrence rate, $h$ is a density for locations, $g$ is the marking density and PP denotes the inhomogeneous Poisson process. We might wish for the density $h$ to capture periodic activity due to time-of-day effects, for example by having the intensity be a step function of the time.

However, real collections of events often exhibit dependencies that cannot be captured by a standard Poisson process (the Poisson process makes the assumption that the number of events that occur in two non-overlapping time intervals must be independent). One way to capture such dependencies is to consider *Cox processes*, which are Poisson processes with a random mean measure. In particular, consider mean measures that take the form of latent Markov processes. In queueing theory, this kind of model is referred to as a Markov-Modulated Poisson Process [Rydén, 1996] and it has been used as a model for packets in networks [Fischer and Meier-Hellstern, 1993].

### 2.1 Events Causing Other Events

In this paper we take a different approach to modeling collections of dependent events in which the occurrence of an event $(t, x)$ triggers a Poisson process consisting of other events. Specifically, we model the triggered Poisson process as having intensity

$$\begin{aligned} k_{(t,x)}(t', x') &= \alpha(x) g_\theta(x'|x) h_\theta(t' - t) \qquad (1) \\ \alpha(x) &\text{ is the expected number of events} \\ h_\theta(t) &\text{ is the delay density} \\ g_\theta &\text{ is the label transition density.} \end{aligned}$$

Denote by $\Pi_0$ the events caused by a baseline Poisson process with mean measure $\mu_0$ and let $\Pi_i$ be the events triggered by events in $\Pi_{i-1}$:

$$\begin{aligned} &D = \cup_i \Pi_i \qquad (2) \\ &\Pi_0 \sim \text{PP}\left(\mu_0\right) \\ &\Pi_i \sim \text{PP}\left(\sum_{(t,x) \in \Pi_{i-1}} k_{t,x}(\cdot, \cdot)\right). \end{aligned}$$

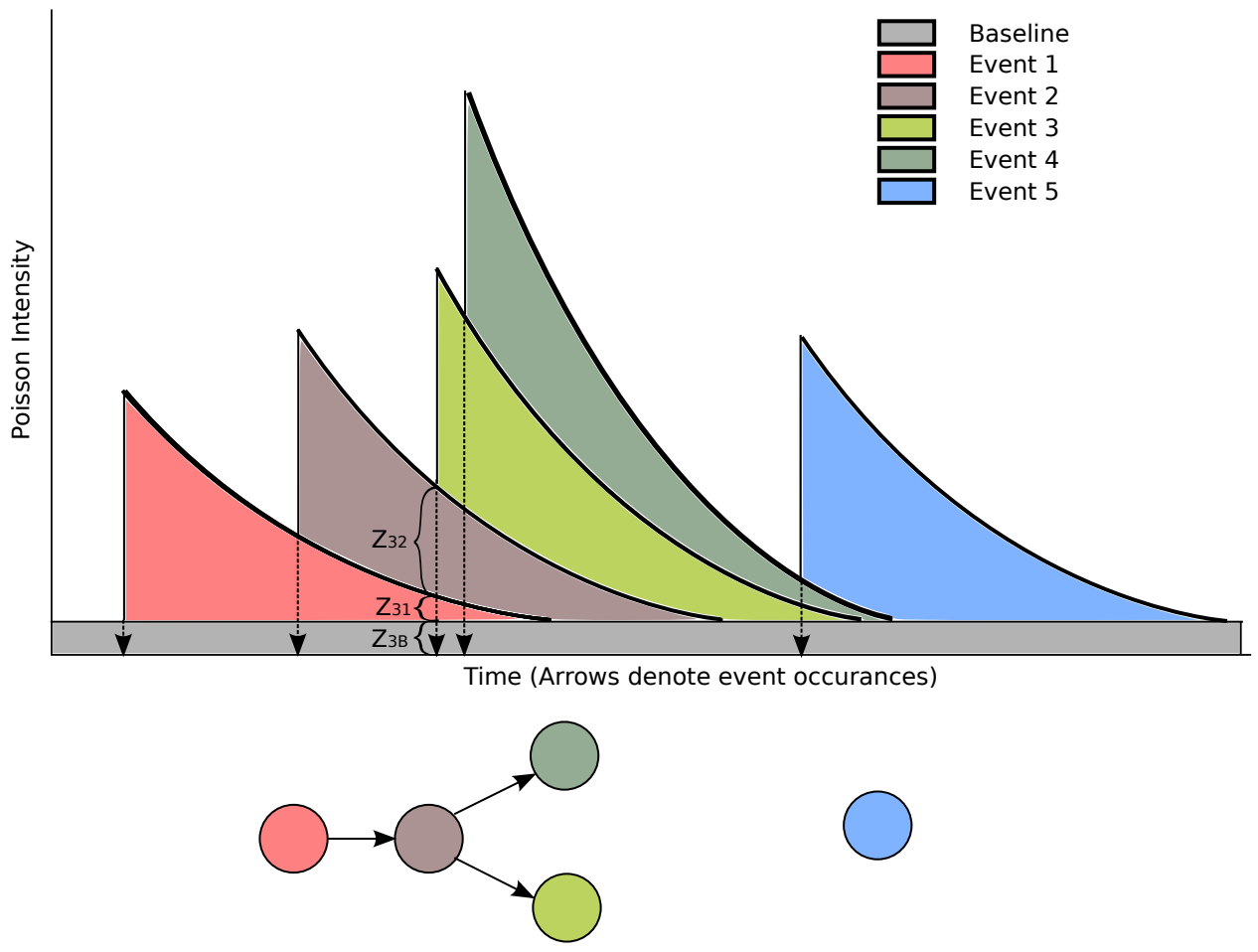

Figure 1: A diagram of the overlapping intensities and one possible forest that corresponds to these events.

Alternatively, we can use the superposition property of Poisson processes to write a recursive definition:

$$D \sim \text{PP}\left(\mu_0 + \sum_{(t,x) \in D} k\left(t', x'\right)\right). \qquad (3)$$

This definition makes sense only when $k_{(t',x')}$ is positive only for $t > t'$, since an event $(t, x)$ can only cause resulting events at a later time, requiring that $h_\theta(t) = 0$ for $t \leq 0$.

**View as a Random Forest** In our model, each event is either caused by the background Poisson process or a previous event (see Figure 1). If we augment the representation to include the cause of each event, the object generated is a random forest, where each event is a node in a tree with timestamp and features attached. The parent of each event is the event that caused it; if that does not exist, it must be a root node.

Let $\pi(p)$ be the event that caused $p$, or $\emptyset$ if the parent does not exist. Usually, this parenthood information is not available and must be estimated, which corresponds to estimating the tree structure from an enumeration of the nodes, their topological sort, timestamps and features. We show how this distribution over $\pi(p)$ can be estimated by an EM algorithm.

### 2.2 Model Fitting

The parameters of our model can be estimated with an EM algorithm [Dempster et al., 1977]. If $\pi(p)$, the cause of the event, was known for every event, then it would be possible to estimate the parameters $\mu_0$, $\alpha$, $g$ and $h$ using standard results for maximum likelihood estimation under a Poisson distribution. Since $\pi$ is not observed, we can use EM to iteratively estimate the latent variables and maximize the parameters. For uniformity of notation, assume that there is a dummy event $(0, \emptyset)$ and $k_{(0,\emptyset)}(t, x) = f_{base}(t, x)$ so that we can treat the baseline intensity the same as all the other intensities resulting from events. We introduce $z_{(t',x',t,x)}$ as expectations of the latent $\pi$ where $z_{(t',x',t,x)}$ corresponds to the expectation of $\mathbf{1}_{(\pi(t,x)=(t',x'))}$. Neglecting terms that don't depend on the EM variables $z$,

$$\begin{aligned}\mathcal{L} &= \sum_{(t,x)\in D} \log \left( \sum_{(t',x')\in D_{0:t}} k_{(t',x')}\left(t, x\right) \right) \\ &\geq \sum_{(t,x)\in D} \left( \sum_{(t',x')\in D_{0:t}} z_{(t',x',t,x)} \log k_{(t',x')}\left(t, x\right) \right) \\ \text{s.t.} & \quad \sum_{t',x'} z_{(t',x,\prime,x,y)} = 1.\end{aligned}$$

The bound is tight when

$$z_{(t',x',t,x)} = \frac{\log k_{(t',x')}\left(t, x\right)}{\sum_{(t',x')} \log k_{(t',x')}\left(t, x\right)}.$$

These $z$ variables act as soft-assignment proxies for $\pi$ and allow us to compute expected sufficient statistics for estimating the parameters in $f_{base}$ and $k$. The specific details of this computation depend on the specific choices made for $f_{base}$ and $k$, but this basically reduces the estimation task to that of estimating a distribution from a set of weighted samples. For example, if $f_{base}(t, x) = \alpha \mathbf{1}_{(0\leq t\leq T)} g(x)$ where $g(x)$ is some labeling distribution, then $\hat{\alpha}_{MLE} = T^{-1} \sum_{(t,x)} z_{(0,\emptyset,t,x)}$.

Regardless of the delay and labeling distributions and the relative intensities of different events, the total intensity of the total mean measure should be equal to the number of events observed. This can either be treated as a constraint during the $M$ step if possible (for example, if $\alpha(x)$ has a simple form), or the results of the $M$ step should be projected onto this set of solutions by scaling $k$ and $f_{base}$, increasing the likelihood in the process.

**Additive components.** It is possible to develop more sophisticated models by making $k_{(t,x)}$ more complex. Consider a mixture $k_{(t,x)}\left(t', x'\right) = \sum_{l=1}^{L} k^{(l)}_{(t,x)}(t', x')$ where $k^{(l)}$ are individual densities. For example, in the Wikipedia edit modeling domain, $k^{(1)}_{(t,x)}$ can produce events similar to $x$ at a time close to $t$, whereas $k^{(2)}_{(t,x)}$ can correspond to more thoughtful responses that occur later but also differ more substantially from the event that caused them. Since the EM algorithm introduces a latent variable for every additive component inside the logarithm, the separation of some components into a further sum can be handled by introducing more latent variables—one for each element. Thus the credit-assigning step builds a distribution not only over the past events that were potential causes, but also the individual components of the mixture.

### 2.3 The Fertility Model

A key design choice is the choice of $\alpha(x)$, the expected number of events. When $x$ ranges over a small space it may be possible to directly estimate $\alpha(x)$ for each $x$. However, with a larger feature space, this approach is infeasible for both computational and statistical reasons and so a functional form of the fertility function must be learned. In presenting these fertility models, we assume for simplicity that $x$ is a binary feature vector.

**Linear Fertility** We consider $\alpha(x) = \alpha_0 + \beta^T x$ with the restriction $\alpha_0 \geq 0, \beta \geq 0$. By Poisson additivity it is possible to factor $\alpha(x)$ into $\alpha_0 + \sum_{i:x_i=1} \beta_i$ and, as part of the EM algorithm, build a distribution over the allocation of features to events, collecting sufficient statistics to estimate the values. Note that $\beta \geq 0$ is an important restriction, since the mean of each of the constituent Poisson random variables must be nonnegative.

This can be somewhat relaxed by considering $\alpha(x) = \alpha_0 + \beta^{+T} x + \beta^{-T}\left(\mathbf{1} - x\right)$ where $\alpha_0 \geq \sum_i \beta_i^-$. Foregoing the $\alpha_0 \geq \sum_i \beta_i^-$ restriction allows the intensity to be negative which does not make probabilistic sense.

**Multiplicative Fertility** The linear model of fertility places significant limits on the negative influence that features are allowed to exhibit and also implies that the fertility effect of any feature will always be the same regardless of its context. Alternatively, we can estimate $\alpha(x) = \exp\left(\beta^T x\right) = \prod_i w_i^{x_i}$ for $w = \exp \beta$, where we assume that one of the dimensions of $x$ is a constant 1, leading to derivatives having the form:

$$\frac{\partial}{\partial w_j}\mathcal{L} = -\sum_{t,x\in D} x_j \prod_{i\neq j} w_i^{x_i} + \sum_{t,x\in D} \sum_{t',x'\in D_{0:t}} z_{(t',x',t,x)} \frac{x_j}{w_j}.$$

The exact solution for a single $w_j$ is readily obtained, so we can optimize $\mathcal{L}$ by either coordinate descent or gradient steps. An alternative approach based on Poisson thinnings is described in Simma [2010].

**Combining Fertilities** It is also possible to build a fertility model that combines additive and multiplicative components:

$$\alpha(x) = \alpha_0^{(0)} + \beta^{(0)T} x + \exp\left(\alpha_0^1 + \beta^{(1)T} x\right) + \cdots.$$

The EM algorithm distributes credit between the constant term $\beta^{(0)T} x$ and the terms $\exp\left(\alpha_0^1 + \beta^{(1)T} x\right)$.

A possible concern is that this requires fitting a large number of parameters. A special case is when $x$ has a particular structure and there is reason to believe that it is composed of groups of variables that interact multiplicatively within the group, but linearly among groups, in which case the multiplicative models can be used on only a subset of variables.

Additionally, it is possible to build a fertility model of the form

$$\alpha(x) = \alpha_0^{(0)} + \beta^{(0)T} x \cdot \exp\left(\alpha_0^1 + \beta^{(1)T} x\right)$$

by using linearity to additively combine intensities and using thinning to handle the multiplicative factors [Simma, 2010].

### 2.4 Computational Efficiency

In this section we briefly consider some of the principal challenges that we needed to face to fit models to massive data (in particular for the Wikipedia data).

For certain selections of delay and transition distributions, it is possible to collapse certain statistics together and significantly reduce the amount of bookkeeping required. Consider a setting in which there are a small number of possible labels, that is, $x_i \in \{1 \dots L\}$ for small $L$, and the delay distribution $h(t)$ is the exponential distribution $h_\lambda(t) = \mathbf{1}_{(\lambda)} \exp(-\lambda x)$. We can use the memorylessness of the exponential distribution to avoid the need to explicitly build a distribution over the possible causes of each event.

Order the events by their times $t_1, \dots, t_n$ and let

$$\begin{aligned} l_{ij} &= \exp\left(\lambda t_{i-1} - \lambda t_i\right) b_{i-1,j} \left(l_{i-1,j} + t_i - t_{i-1}\right) / b_{ij} \\ b_{ij} &= \exp\left(\lambda t_{i-1} - \lambda t_i\right) b_{i-1,j} + \alpha(x_i) g(j|x_i). \end{aligned}$$

Let $i(s) = \inf\{t_i : t_i < s\}$ and note that the intensity at time $s$ for a label of type $j$ is

$$\exp\left(\lambda t_{i(s)} - \lambda s\right) b_{i(s),j} + f_{base}(s, j),$$

and the weighted-average delay is $l_{i(s),j} + s - t_{i(s)}$. Counting the number of type $j$ events triggering type $k$ can be done with similar techniques by letting $b_{i,j,k}$ (the intensity at time $i(s)$ for events $j$ caused by $k$) change only when an event $k$ is encountered. If the transition density is sparse, only some $b_{ij}$ need to be incremented and the rest may be left unmodified, as long as the missing exponential decay is accounted for later. While this computational technique works for only a restricted set of models and has computational complexity $O(|D|\bar{z})$ where $\bar{z}$ is the average number of non-zero $k(\cdot, x)$ entries, it is much more computationally efficient than the direct method when there are a large number of somewhat closely spaced events.

For large-scale experiments on Wikipedia, we use Hadoop, an open-source implementation of MapReduce [Dean and Ghemawat, 2004]. The object that we map over is a collection of a page and its neighbors in the link graph.[1] Each map operation also accesses the hyperparameters shared across pages and runs multiple EM iterations over the events associated with that page. The learned parameters are returned to the reducer which updates the hyperparameters and another MapReduce job fits models with these updated hyperparameters. Thus, the reduce step only accumulates statistics for the hyperparameters, as well as collects log-likelihoods.

Hadoop requires that each object being mapped over be kept in memory, which requires careful attention to representation and compression; these memory limits have been the key challenge in scaling. If each neighborhood does not fit in memory, it is possible to break it into pieces, run the E step in the Map phase and then use the Reduce phase to sum up all the sufficient statistics and maximize parameters, but this requires many more chained MapReduce jobs, which is inefficient. For our experiments, careful engineering and compression was sufficient.

## 3 Twitter Messages

Twitter is a popular microblogging website that is used to quickly post short comments for the world to see. We collected Twitter messages (composed of the sender, timestamp and body) that contained references to stock tickers in the message body. Some messages form a conversation; others are posted as a result of a real-world event inspiring the commentary. The dataset that we collected contains 54717 messages and covers a period of 39 days. For modeling, each message can be represented as a triple of a user, timestamp and a binary vector of features. A typical message

```
User:  SchwartzNow
Time:  2009-12-17T19:20:15
Body:  also for tommorow expect
high volume options traded stocks
like $aapl,$goog graviate around the
strikes due to the delta hedging
```

[1] This is generated with a sequence of MapReduce jobs where we first compute diffs and featurize, then for each page we gather a list of neighbors that require that page's history, and finally each page sends a copy of itself to all its neighbors. A page's body is insufficient to determine its neighbors since the body only contains outgoing (not incoming) links so the incoming links need to be collected first.

occurs on 2009-12-17 at 19:20:15 and has the features $AAPL and $GOOG and is missing features such as $MSFT and HAS_LINK. Due to length constraints and Internet culture, the messages tend to not be completely grammatical English and often a message is simply a shortened Web link with brief commentary. In addition to the stocks involved and whether links are involved, features also denote the presence or absence of keywords such as "buy" or "option."

**Baseline Intensities** The simplest possible baseline intensity is a time-homogeneous Poisson process, but the empirical intensity is very periodic. A better baseline is to break up the day into intervals of (for example) an hour, assume that the intensity is uniform within the hour and that the pattern repeats. So, $h(t) = p_{\lfloor t/24 \rfloor}$. The log-likelihoods for these baselines are reported in Table 1. It is worth noting that the gain from incorporating periodicity in the baseline is much smaller than the gain from the other parts of the model.

This timing model must be combined with a feature distribution. We use a fully independent model, where each feature is present independently of the others. That is, $g(x) = \prod_i p_i^{g_i(x)} (1-p_i)^{1-g_i(x)}$, where $g_i$ is the $i^{th}$ feature. Clearly, the MLE estimates for $p_i$ are simply the empirical fraction of the data that contains that feature.

### 3.1 Intensity and Delay Distributions

When events can trigger other events, each induces a Poisson process of successor events. We factor the intensity for that process as $k_{(t,x)}(t', x') = \alpha(x) g(x'|x) h(t'-t)$, with the constituents described in Eq. 1. For the intensity, we implemented a multiplicative model where the expected number of events is $\alpha(x) = \exp(\beta^T x)$. The delay distribution $h$ must capture the empirical fact that most responses occur shortly after the original message, but there exist some responses that take significantly longer, meaning that $h$ needs a sufficiently heavy tail. As candidates, we consider uniform, piecewise uniform, exponential and gamma distributions.

Log-likelihoods for different delays are reported in Figure 2. The transition function used, $g_\gamma$, is described later. The best performing delay distribution is the gamma, with shape parameters less than 1; the shape parameter is also estimated in the results of Table 1. Note that the results show that the choice of a delay distribution has a smaller impact on the overall likelihood than the transition distribution. This is due in part to the fact that for an individual event the features are embedded in a large space and there is more to explain. The predictive ability of the Poisson process associated with an event to explain the specific features of a resultant event is the predominant benefit of the model.

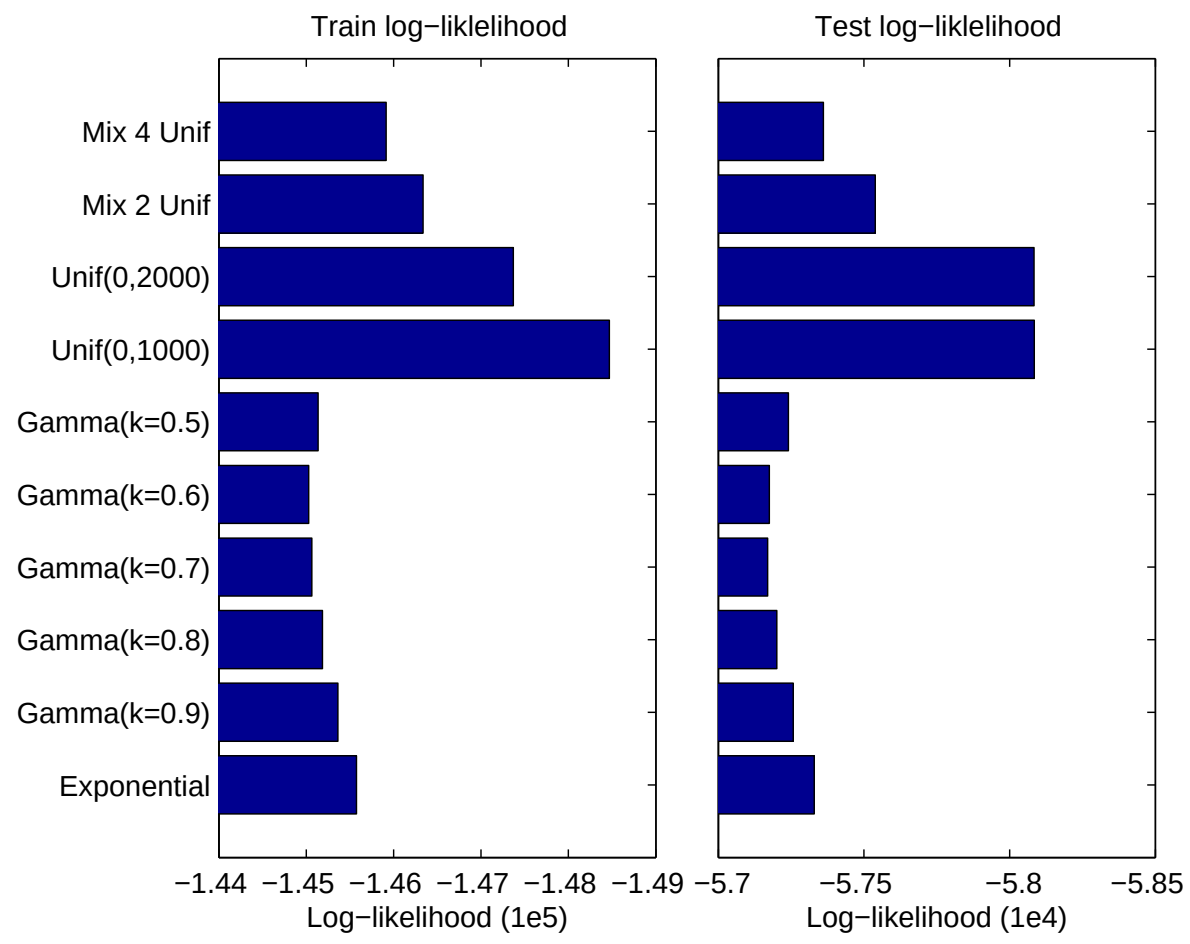

Figure 2: Log-likelihoods for various delay functions.

### 3.2 Transition Distribution

The remaining aspect of the model is the transition distribution $g(x|x')$ that specifies the types of events that are expected to result from an event of type $x'$. Let's consider the possible relationships between a message and its trigger:

1. A simple 'retweet'—a duplication of the original message.
2. A response—a message either prompts a specific response to the content of the message, or motivates another message on a similar topic.
3. After a message, the probability of another (possibly unrelated) message is increased because the original event acts as a proxy for general user activity. These kinds of messages represent variation in the baseline event rate not captured by the baseline process and are unrelated to the triggering message in content, so they should take on a distribution from the prior.

We construct a transition function parametrized by $\gamma$ that is a product of independent per-feature transitions, each a mixture of the identity function and the prior:

$$g_\gamma(x, x') = \prod_i \left( (1-\gamma)\, \mathbf{1}_{(x_i = x'_i)} + \gamma p_i^{x'_i} \left(1 - p_i^{1-x'_i}\right) \right).$$

Note that $g_\gamma$ is *not* a mixture of the identity and the prior.

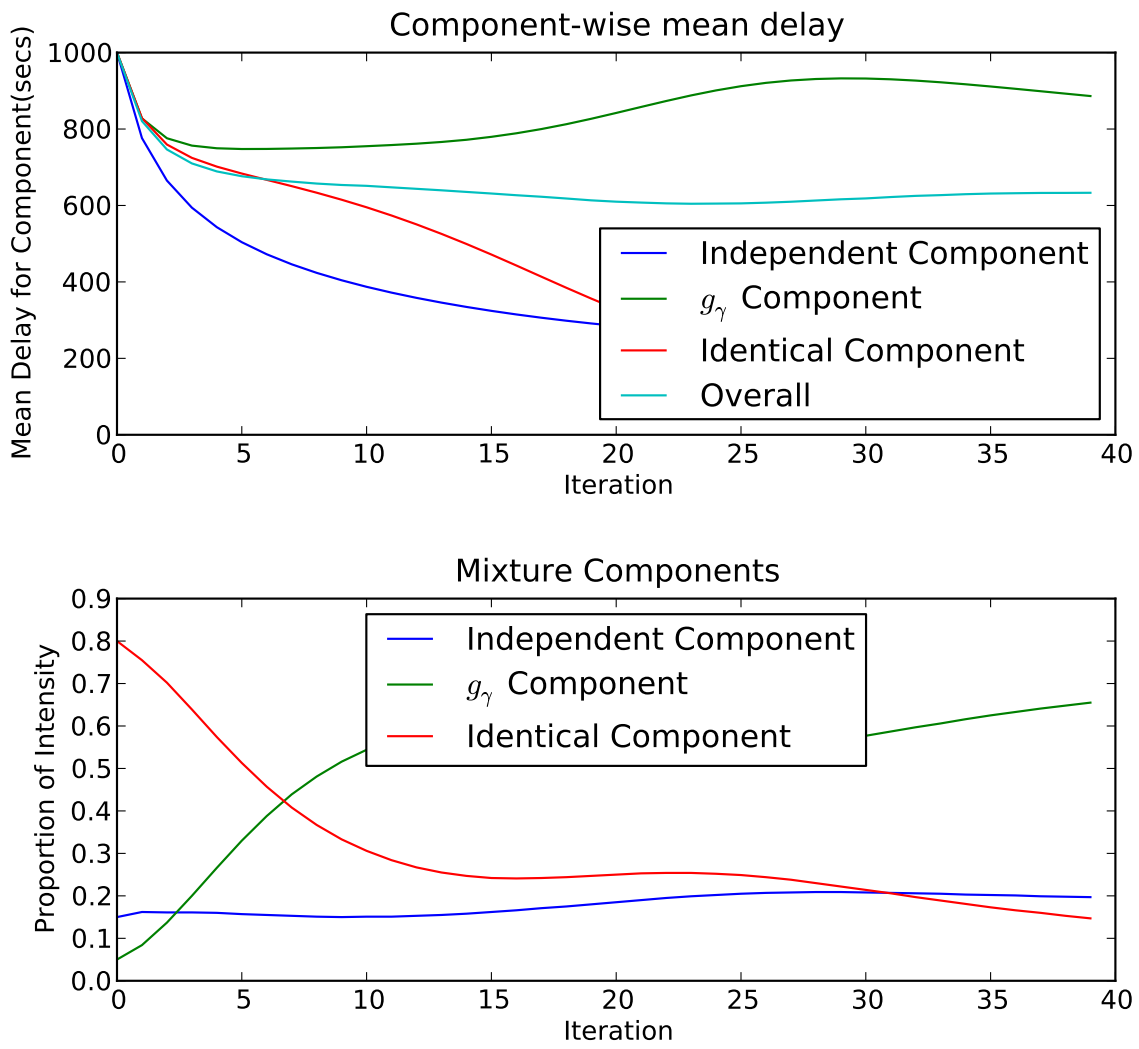

Figure 3: Trace of parameters of the individual mixture components in model 5.

Table 1: Log-likelihoods for models of increasing sophistication.

| Type | Train | Test |
|---|---|---|
| Homogeneous Baseline Only | -167810 | -66050 |
| Periodic Baseline Only | -164695 | -64758 |
| Exp Delay, Independent transition($k_1$) | -161905 | -63017 |
| Intensity doesn't depend on features, Exp Delay, $g_\gamma$ transition | -145752 | -57383 |
| Feature-dependent intensity, Exp Delay, Identity transition ($k_3$) | -146558 | -57810 |
| Exp Delay, $h_\gamma$ transition ($k_2$) | -145557 | -57313 |
| Shared intensity, shared Exp delay, mixture transition ($k_4$) | -145629 | -57379 |
| Mixture of (intensity, exp delay, different transitions) ($k_5$) | -145152 | -57130 |
| Mixture of (intensity, gamma delay, different transitions) | -144621 | -56966 |

We denote two important special cases as $g_1$, where each resultant event is drawn independently, and $g_0$, where the caused events must be identical to the trigger. With an exponential delay distribution and $\alpha(x)$ fixed at 1, $g_0$ is equivalent to setting the Poisson intensity to an exponential moving average with decay parameter determined by $\lambda$. The EM algorithm can be used to find the optimal decay parameter, but as the reported results show, this model is inferior to one that utilizes the features of the events.

Earlier, we enumerated relationships between a message and its trigger. For example, the retweets are completely identical to the original, with the possible exception of a "@username" reference tag, so the transition would be $g_0$. A response would have similar features but may differ in a few features, and a density-proxy message would have features independent of the causing message, corresponding to $g_\gamma$ for $0 < \gamma < 1$. $g_1$ models the density-proxy phenomenon.

Let us now consider some possible models, where the Greek letters represent parameters to be estimated:

$$
\begin{aligned}
k_{1(t,x)}(t',x') &= \exp\left(\alpha_1 + \beta_1^T x\right) h_1\left(t'-t\right) g_1(x,x') \\
k_{2(t,x)}(t',x') &= \exp\left(\alpha_2 + \beta_2^T x\right) h_2\left(t'-t\right) g_\gamma(x,x') \\
k_{3(t,x)}(t',x') &= \exp\left(\alpha_3 + \beta_3^T x\right) h_3\left(t'-t\right) g_0(x,x') \\
k_{4(t,x)}(t',x') &= \exp\left(\alpha_4 + \beta_4^T x\right) h_4\left(t'-t\right) \times \\
&\quad (\eta_1 g_1(x,x') + \eta_2 g_\gamma(x,x') + \eta_3 g_0(x,x')) \\
k_{5(t,x)}(t',x') &= \sum_{i=1}^{3} k_{i(t,x)}(t',x').
\end{aligned}
$$

The models $k_i$ for $i$ from 1 to 3 are designed to capture the $i^{th}$ phenomenon, while $k_4$ and $k_5$ are intended to capture all three effects. Both $g$ and $h$ are densities, so it's easy to compute $\int k_{(t,x)}(t,x,t',x')dt'dx'$. The results, shown in Figure 1, indicate that models 4 and 5 are significantly superior to the first three, demonstrating that separating the multiple phenomena is useful. For $h$, we use an exponential distribution.

In model 4, all the transition distributions share the same fertility and delay functions,whereas in model 5, each distribution has its own fertility and delay. As shown in Figure 3, the latter performs significantly better, indicating that the three different categories of message relationships have different associated fertility parametrizations and delays. The top plot shows the proportions of each component in the mixture, defined as the ratio of the average fertility of the component to the total fertility. The bottom plot demonstrates that while the mean delay of the overall mixture remains almost constant throughout the EM iterations, different individual components have substantially different delay means.

### 3.3 Results and Discussion

Table 1 reports the results for a cascade of models of increasing sophistication, demonstrating the gains that result from building up to the final model. The first stage of improvements, from the homogeneous to the periodic baseline and then to the independent transition model focuses on the times at which the events occur, and shows that roughly equivalent gains follow from modeling periodicity and from further capturing less periodic variability with an exponential moving average. The big boost comes from a better labeling distribution that allows the features of events to depend on the previous events, capturing both the topic-wise hot trends and specific conversations.

Of course, the shape of the induced Poisson process has an effect. The different types of transitions have distinctly different estimated means for their delay distributions, which is to be expected since they capture different effects. As seen in Figure 3 the overall-intensity proxying independent transition has the highest mean, since the level of activity, averaged over labels, changes slower than the activity for a particular stock or topic. For shape, lower $k$, higher-variance gamma distributions work best.

The final component is a fertility model that depends on the features of the event and allows some events to cause more successors than others. This actually has less impact on the log-likelihood than the other components of the model.

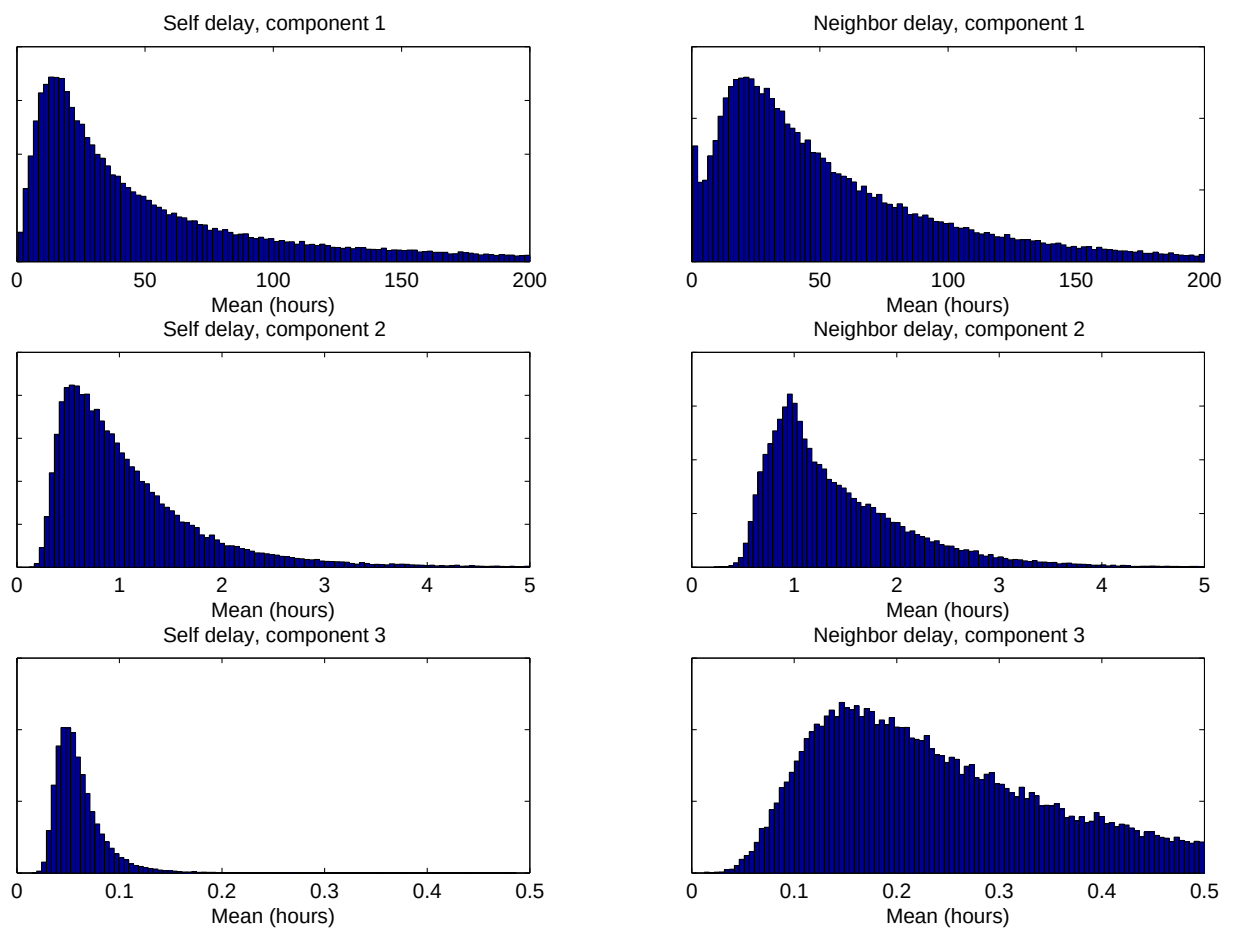

Figure 4: Delay distribution histogram over all pages.

# 4 Wikipedia

Wikipedia is a public website that aims to build a complete encyclopedia through user edits. We work to build a probabilistic model for predicting edits to a page based on revisions of the pages linking to it. Causes outside of that neighborhood are not considered. The reasons for that restriction are primarily computational—considering all edits as potential causes for all other edits, even within a short time window, is impractical on such a large scale. As a demonstration of scale, we model 414,540 pages with a total of 71,073,739 revisions (the raw datafile is 2.8TB in size), involving billions of considered interactions between events.

## 4.1 Structure in Wikipedia's History

As we build up a probabilistic model for edits, it's useful to consider the kinds of structure we would like the model to capture. Edits can be broadly categorized into:

**Minor Fixes**: small tweaks that include spelling corrections, link insertion, etc. Only one or a few words in the document are affected.

**Major Insert**: Often, text is migrated from a different page such that we obtain the addition of many words and the removal of none or very few. From the user's perspective, this corresponds to typing or pasting in a body of text with minimal editing of the context.

**Major Delete:** The opposite of a major insert. Often performed by vandals who delete a large section of the page.

**Major Change**: An edit that affects a significant number of words but is not a simple insert or delete.

**Revert:** Any edit that reverts the content of the page to a previous state. Often, this is the immediately previous state but sometimes it goes further back. A revert is typically a response to vandalism, though edits done in good faith can also be reverted.

**Other Edit**: A change that affects more than a couple of words but is not a major insert or delete.

## 4.2 Delay Distributions

Since most pages have many neighbors, each event has a large number of possible causes and the mean measure at each event is the sum over many possible triggers. This means the exact shape of the delay distribution is not as important as in cases when only a few possible triggers are considered. We model the delay as a mixture of three exponentials, intending them to capture short, medium and longer-term effects. For each page, we estimate both the parameters and the mixing weights. Figure 4 shows a histogram of the estimated means.

One component is a very fast response, with an average of 3.6 minutes for the same-page and 13.8 minutes for the adjacent-page delay. On the same page, the component captures edits caused by each other, either when an individual is making multiple modifications and saving the page along the way, or when a different user noticing the revisions on a news feed and instantly responding by changing or undoing them. The remaining components capture the periodic effects and time-varying levels of interest in the topic, as well as reactions to specific edits.

### 4.3 Transition Distribution

The model needs to capture the significant attributes of the revision, in addition to its timestamp, but we don't aim to completely model the exact content of the edit, as the inadequacies of that aspect of the model would dominate the likelihood. Instead, we identify key features (type—revert, major insert, etc—whether the edit was made by a known user, and the identity of the page) of the edits and build a distribution over events as described by those features, not the raw edits.

When a page with features $x$ triggers an event with features $x'$, the latter vector is drawn from a distribution over possible features. When the number of possible feature combinations is small, the transition matrix can be directly learned, but when there are multiple features, or features which can take on many values, we need to fit a structured distribution. We partition the features into two parts as $x = (x_1, x_2)$, where $x_1$ are features that can appear in any revision (such as the type of the edit and whether the editor is anonymous) and where $x_2$ is the identity of the page. Note that $x_2$ can take on very many values, each one appearing relatively infrequently. There are a vast number of observations and we can directly learn the transition matrix $h_1(x_1, x_1')$. For each target page $x_2'$, we model an $x_1$ transition as

$$\begin{aligned} x_1'|x_1, x_2 &\sim \text{Multinomial}\left(\theta_{x_1,x_2}\right) \\ \theta_{x_1,x_2} &\sim \text{Dirichlet}\left(\gamma_{x_1}\right) \end{aligned}$$

which, due to conjugacy, corresponds to shrinkage towards $\gamma_{x_1}$. As more transitions are observed, the page's transition probability becomes more driven by the specific observed probabilities on that page. The allocation over components of $\gamma$ is directly maximized, while the magnitude of $\gamma$ is chosen over a validation set. $x_2$ is handled by fixing a particular page that we refer to as $x_2^\star$ and fitting a model for revisions of that page, $(x_1, x_2^\star)$. Then, the process over all the pages is a superposition of processes over each possible $x_2$.

Figure 5 shows log-likelihoods of successive iterations of the model. The regularized versions use the Dirichlet prior; the others estimate $\theta$ on each page independently. The bars correspond to:

- **No Neighbors**: The revisions on each page can be caused either by the baseline or a previous revision on that page but not by revisions of the neighbors:

$$k^{x_2^\star}_{(t,x)}\left(t', x'\right) = \mathbf{1}_{\left(x_2 = x_2^\star\right)} \alpha g(x'|x) h(x, x', t'-t).$$

- **Neighbors, Same Transition**: Revisions to the neighbors of the page in the link graph cause a Poisson process of edits on the page. That process has its own delay distribution and intensity, but those are the same for all neighbors. The transition conditional distribution is the same for both events

$$\begin{aligned} k^{x_2^\star}_{(t,x)}\left(t', x'\right) &= \mathbf{1}_{\left(x_2 = x_2^\star\right)} \alpha_s g(x'|x) h_s(t'-t) \\ &+ \mathbf{1}_{\left(x_2 \in \delta x_2^\star\right)} \alpha_n g(x'|x) h_n(t'-t). \end{aligned}$$

  Parameters for functions with different subscripts are estimated separately.

- **Neighbors, Different Transitions**: Same as above, but uses different transition distributions for $x_2^\star$ and its neighbors:

$$\begin{aligned} k^{x_2^\star}_{(t,x)}\left(t', x'\right) &= \mathbf{1}_{\left(x_2 = x_2^\star\right)} \alpha_s g_s(x'|x) h_s(t'-t) \\ &+ \mathbf{1}_{\left(x_2 \in \delta x_2^\star\right)} \alpha_n g_n(x'|x) h_n(t'-t). \end{aligned}$$

  Here, the parameters for the two different $g$ are estimated separately and are regularized towards $\gamma$ same or $\gamma$ neighbor, respectively.

- **Neighbors, Own Intensities**: Each neighbor has its own $\alpha$ parameter:

$$\alpha(x, x') = \mathbf{1}_{\left(x_2^\star = x_2', x_2 \text{neighbor of } x_2^\star\right)} \alpha_{x_2}.$$

  For most pages there is insufficient data to estimate the individual $\alpha$s accurately; regularization of $\alpha$ is required and is discussed later.

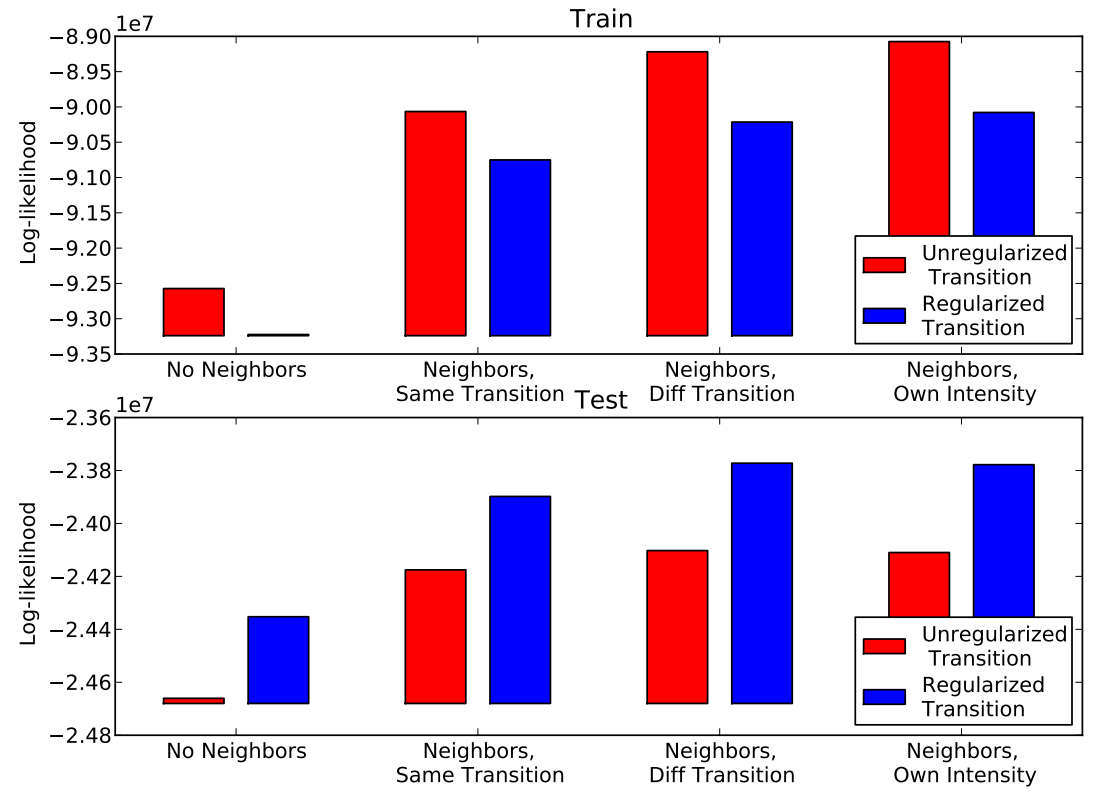

Figure 5: Log-Likelihoods of various models. Models with regularized transition matrices perform significantly better on unseen data, but non-trivially worse on the training set, indicating strong regularization. The baseline-only is not shown but has $-1.48 \times 10^8$ training and $-3.98 \times 10^7$ test log-likelihoods.

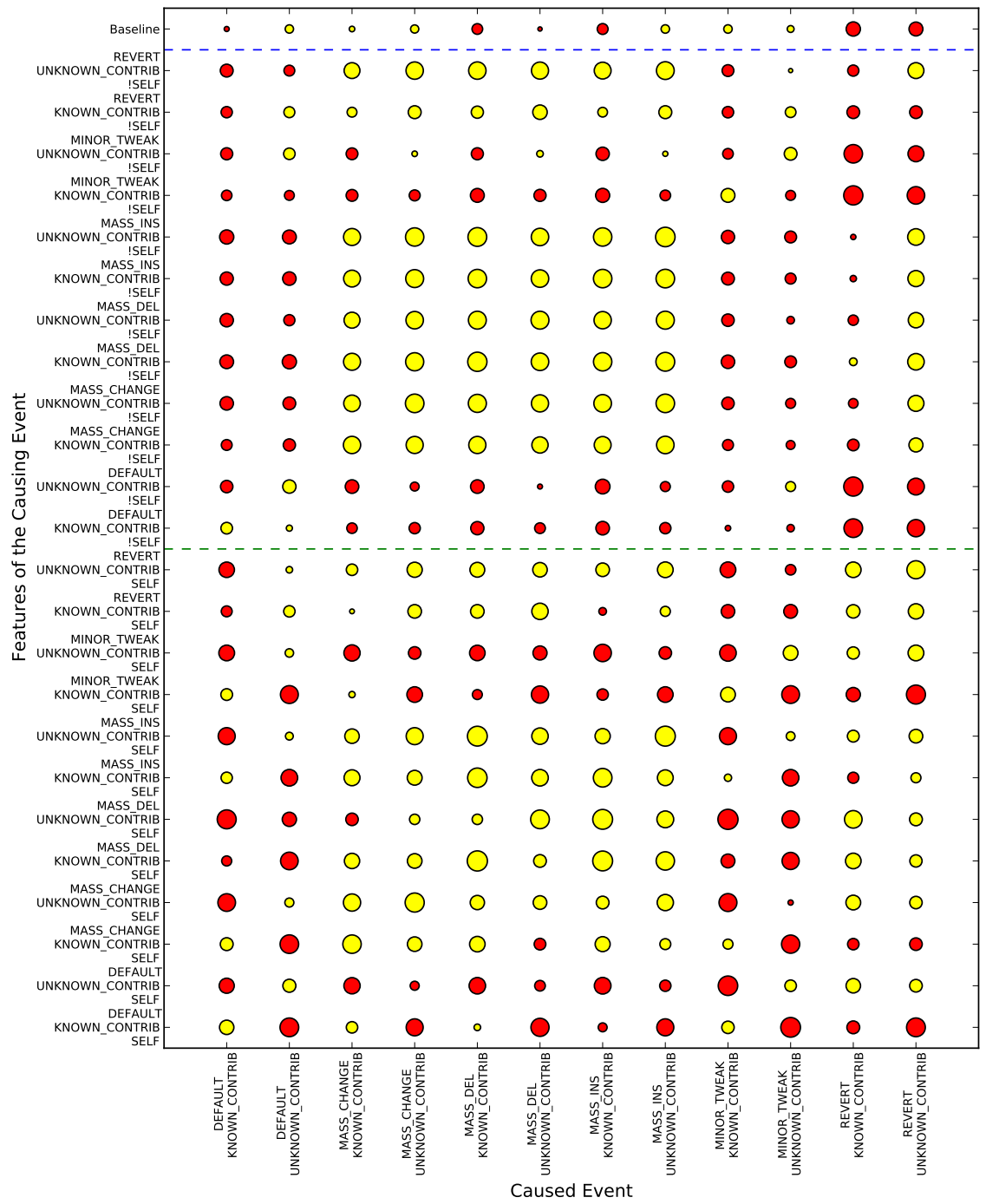

Figure 6: Learned Transition Matrix. The area of the circles corresponds to the logarithm of the conditional probability of the observed feature, divided by the marginal. The yellow, light-colored circles correspond to the transition being more likely than average; red correspond to the transition being less likely.

### 4.4 Learned Transition Matrices

Figure 6 shows the estimated transition matrix. Each circle denotes $\log(g(x, x')/p(x'))$; when it is high, that label of the caused event is much more likely than it would be otherwise.

The top row represents the intensity for the baseline, the labels of events whose cause is not a previous event. Positive values correspond to event types that the events-triggering-events aspect of the model is less effective in capturing and thus are over-represented in the otherwise-unexplained column. Reverts, both by known and anonymous contributors, are significantly underrepresented, indicating that the rest of the model is effective in capturing them. Revisions made by known contributors are under-represented, as the rest of the model captures them better than the edits made by anonymous contributors. Events generated from this row account for 23.87% of total observed events.

The next block corresponds to edits on neighbors causing revisions of the page under consideration and are responsible for 19.11% of observed events. The diagonal is predominantly positive, indicating that an event of a particular type on a neighbor makes an event of the same type more likely on the current page. Note the significantly positive rectangle for transitions between massive inserts, deletions and changes. The magnitude of the ratio is almost identical in the rectangle; significant modifications induce other large modifications but the specific type of modification, or whether it is made by a known user, are irrelevant. Large changes act as indications of interest in the topic or significant structural changes in the related pages.

The remaining block represents edits on a page causing further changes on the same page and is responsible for 57.02% of the observations. There is a stronger positive diagonal component here than above, as similar events co-occur. Large changes, especially by anonymous users, lead to an over-representation of reverts following them. On the other hand, reverts result in extra large changes, as large modifications are made, reverted and come back again feeding an edit war. Reverts actually over-produce reverts. This is not a first-order effect, since reverts rarely undo the previous undo, but rather captures controversial moments. The presence of a revert is an indication that previously, an unmeritorious edit was made, which suggests that future unmeritorious edits (that tend to be long and spammy) that need to be reverted are likely.

### 4.5 Regularizing Intensity Estimates

When for a fixed page $x_2^\star$ an edit occurs on its neighbor, one would expect the identity of the neighbor to affect its likelihood of causing an event on $x_2^\star$. As it turns out, effectively estimating the intensities between a pair of pages is impractical unless a very large number of revisions have been observed. Even in the high-data regimes, strong regularization is required. We tried regularizing fertilities both towards zero and toward a common per-page mean, using both $L_1$ and $L_2$ penalties, but these regularizers empirically led to poorer likelihoods than using a single scalar $\alpha$ for all neighbors, suggesting that there is not enough data to accurately estimate individual $\alpha$s. One reason is that pages with a large number of events also have a large number of neighbors, so the estimation is always in a difficult regime. Furthermore, the hypothetical 'true' values of these parameters will change with time, as new neighbors appear and change.

Let $m_i$ be the number of revisions of the $i^{th}$ neighbor page and let $n_i$ be the expected number of events triggered by that neighbor's revisions. One approach that works in high-data regimes is to let

$$\hat{\alpha}_{i,REG} = \lambda \frac{\sum_j n_j}{\sum_j m_j} + (1 - \lambda)\, \frac{n_i}{m_i},$$

Table 2: Sample list of pages (in bold) and the intensities estimated for them and their top neighbors. This is under strong regularization, which explains the similarity of the weights.

| Page | Int. | Page | Int. |
|---|---|---|---|
| **AH-64 Apache** | **0.49** | **South Pole** | **0.46** |
| AH-1 Cobra | 0.063 | Equator | 0.017 |
| CH-47 Chinook | 0.040 | Roald Amundsen | 0.016 |
| 101st Airborne Division | 0.040 | Ernest Shackleton | 0.016 |
| Mil Mi-24 | 0.037 | Geography of Norway | 0.015 |
| Flight simulator | 0.037 | Navigation | 0.015 |
| List of Decepticons | 0.034 | South Georgia and the South Sandwich Islands | 0.014 |
| Tom Clancy's Ghost Recon Advanced Warfighter | 0.034 | National Geographic Society | 0.014 |
| Command & Conquer | 0.033 | List of cities by latitude | 0.014 |

for a parameter $\lambda$ between zero and one, which yields an average between the aggregate and individual maximizers. The regularizer forces the lower weights to clump as each is lower-bounded by $\lambda \sum n_j / \sum m_j$. On a subset of the Wikipedia graph that includes only pages with more than 500 revisions, this improves held-out likelihoods compared to having a single $\alpha$ for all neighbors. The improvement is very small, however, certainly smaller than the impact of other aspects of the model. Example pages and intensities estimated for their neighbors are shown in Table 2.

## 5 Conclusions

We have presented a framework for building models of events based on cascades of Poisson processes, demonstrated their applications and demonstrated scalability on a massive dataset. The techniques described in this paper can exploit a wide range of delay, transition and fertility distributions, allowing for applications to many different domains.

One direction for further investigation is to provide support for latent events that are root causes for some of the observed data. Another is a Bayesian formulation that integrates instead of maximizes parameters; this may work better for complex fertility or transition distributions that lack sufficient observations to be accurately fit with maximum likelihood. Both extensions complicate inference and reduce scalability; indeed, Wingate et al. [2009] propose a Bayesian model with latent events but scaling is an issue. Furthermore, allowing the parameters of the model to depend on time (for example, letting the fertility be a draw from a Gaussian process) would be very useful, though again, computational issues are a concern.

## 6 Acknowledgements

We gratefully acknowledge support for this research from Google, Intel, Microsoft and SAP.

## References

J. Dean and S. Ghemawat. MapReduce: simplified data processing on large clusters. In *Symposium on Operating Systems Design & Implementation (OSDI)*, 2004.

A. P. Dempster, N. M. Laird, and D. B. Rubin. Maximum likelihood from incomplete data via the EM algorithm. *Journal of the Royal Statistical Society. Series B (Methodological)*, 39(1):1–38, 1977.

W. Fischer and K. Meier-Hellstern. The Markov-modulated Poisson process (MMPP) cookbook. *Performance Evaluation*, 18:149–171, 1993.

A. G. Hawkes. Spectra of some self-exciting and mutually exciting point processes. *Biometrika*, 58(1): 83, 1971.

U. Nodelman, C. R. Shelton, and D. Koller. Continuous time Bayesian networks. In *Uncertainty in Artificial Intelligence (UAI)*, 2002.

U. Nodelman, C. R. Shelton, and D. Koller. Expectation maximization and complex duration distributions for continuous time Bayesian networks. In *Uncertainty in Artificial Intelligence (UAI)*, 2005.

Y. Ogata. Statistical models for earthquake occurrences and residual analysis for point processes. *Journal of the American Statistical Association*, 83 (401):9–27, 1988.

S. Rajaram, T. Graepel, and R. Herbrich. Poisson-networks: A model for structured point processes. In *International Workshop on Artificial Intelligence and Statistics (AISTAT)*, 2005.

T. Rydén. An EM algorithm for estimation in Markov-modulated Poisson processes. *Computational Statistics and Data Analysis*, 21:431–447, 1996.

A. Simma. *Modeling Events in Time using Cascades of Poisson Processes.* PhD thesis, University of California, Berkeley, 2010.

A. Simma, M. Goldszmidt, J. MacCormick, P. Barham, R. Black, R. Isaacs, and R. Mortier. CT-NOR: Representing and reasoning about events in continuous time. In *Uncertainty in Artificial Intelligence (UAI)*, 2008.

D. Wingate, N. D. Goodman, D. M. Roy, and J. B. Tenenbaum. The infinite latent events model. In *Uncertainty in Artificial Intelligence (UAI)*, 2009.